\documentclass{article}

\usepackage[final]{neurips_2025}

\usepackage[utf8]{inputenc} 
\usepackage[T1]{fontenc}    
\usepackage{hyperref}       
\usepackage{url}            
\usepackage{booktabs}       
\usepackage{amsfonts}       
\usepackage{algorithm}
\usepackage{algpseudocode}
\usepackage{wrapfig}
\usepackage{amsmath}
\usepackage{amssymb}
\usepackage{bm}
\usepackage{svg}
\usepackage{nicefrac}       
\usepackage{microtype}      
\usepackage{xcolor}         
\usepackage{multirow}
\usepackage{graphicx}
\usepackage{subcaption}
\usepackage{cleveref}
\usepackage{fancyvrb}
\usepackage{xcolor}
\usepackage{algorithm}
\usepackage{algpseudocode}
\usepackage{amsmath}

\title{Symphony for Medical Coding: \\A Next-Generation Agentic System for Scalable and Explainable Medical Coding}

\author{%
  Joakim Edin$^{\dagger}$ \\
  \And
  Andreas Motzfeldt$^{\dagger}$ \\
  \And
  Simon Flachs \\
  \And
  Lars Maal{\o}e* \\
  \AND
  Corti \\
}

\begin{document}

\maketitle
\begin{abstract}
    Medical coding translates free-text clinical documentation into standardized codes drawn from classification systems that contain tens of thousands of entries and are updated annually. It is central to billing, clinical research, and quality reporting, yet remains largely manual, slow, and error-prone. Existing automated approaches learn to predict a fixed set of codes from labeled data, thereby preventing adaptation to new codes or different coding systems without retraining on different data. They also provide no explanation for their predictions, limiting trust in safety-critical settings. We introduce Symphony for Medical Coding, a system that approaches the task the way expert human coders do: by reasoning over the clinical narrative with direct access to the coding guidelines. This design allows Symphony to operate across any coding system and to provide span-level evidence linking each predicted code to the text that supports it. We evaluate on two public benchmarks and three real-world datasets spanning inpatient, outpatient, emergency, and subspecialty settings across the United States and the United Kingdom. Symphony achieves state-of-the-art results across all settings, establishing itself as a flexible, deployment-ready foundation for automated clinical coding.
\end{abstract}

\section{Introduction}
\begin{wrapfigure}{r}{0.45\linewidth}
\vspace{-0.6cm}
    \centering
    \includegraphics[width=\linewidth]{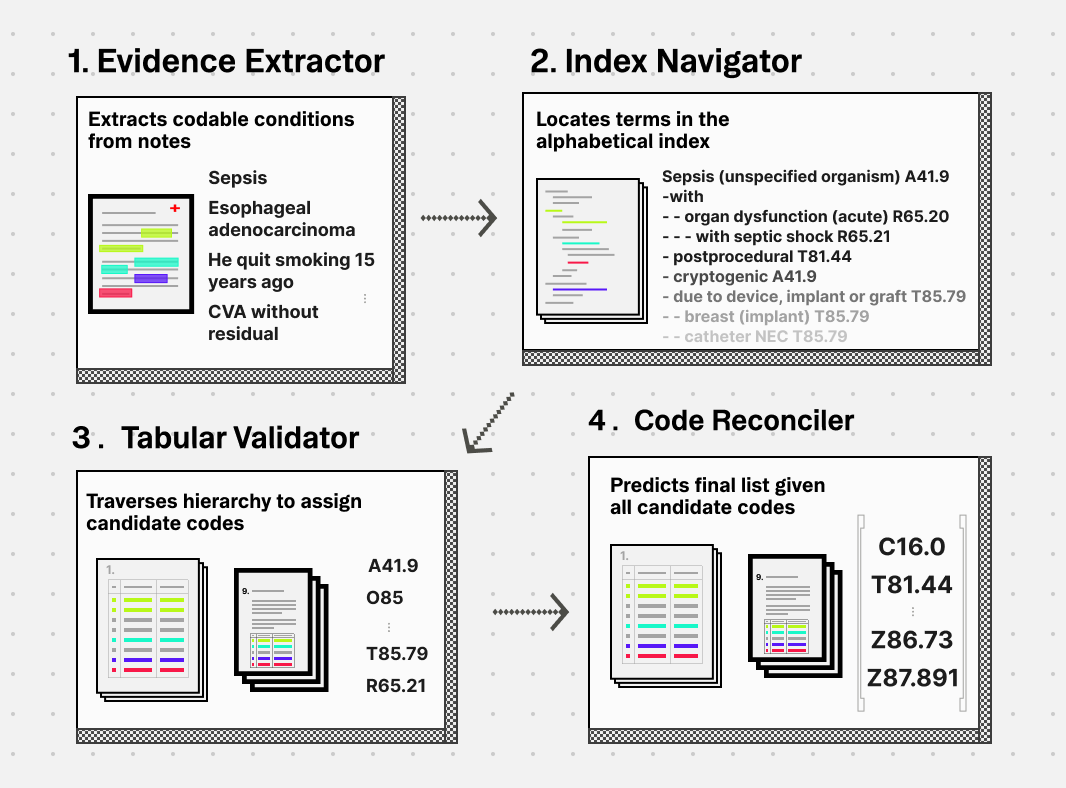}
    \caption{Overview of the reasoning workflow learned in Symphony for Medical Coding.}
    \vspace{-0.5cm}
\end{wrapfigure} In August 2025, the Corti team introduced Code Like Humans (CLH), a fundamentally different approach to automated medical coding \citep{motzfeldtCodeHumansMultiAgent2025}. Rather than treating coding as a purely supervised multi-label classification problem learned from a fixed dataset \citep{huangPLMICDAutomaticICD2022,edinAutomatedMedicalCoding2023a}, CLH decomposed the task into a structured multi-agent workflow that emulates the reasoning process of expert human coders with full access to reference materials. This paradigm shifted the focus from closed-set statistical prediction to ontology-aware, reasoning-driven code assignment. It has since inspired a new class of agentic coding systems, with teams at Oracle Health \& AI and AWS AI achieving state-of-the-art performance using approaches based on large language models such as OpenAI's GPT and Anthropic's Claude \citep{zhengMedDCRLearningDesign2025, yuanReliableClinicalCoding2025}.

We now present \textbf{Symphony for Medical Coding}, a next-generation medical coding system that advances both performance and deployment readiness. Symphony achieves state-of-the-art accuracy while preserving the flexibility to work with any coding ontology, even as it changes.

In this paper, we make the following contributions:

\begin{enumerate}
    \item \textbf{Benchmark performance:} We demonstrate state-of-the-art results on major public medical coding benchmarks, outperforming leading foundation models and healthcare software providers.
    
    \item \textbf{Real-world validation:} We report superior performance on production data from a large U.S. and U.K. provider system, highlighting robustness beyond curated benchmark datasets.
    
    \item \textbf{Subspecialty adaptation:} We show that Symphony can be systematically adapted to subspecialty-specific contexts, yielding further performance improvements without retraining.
    
    \item \textbf{High-precision operation:} We demonstrate that the system maintains state-of-the-art precision, a critical requirement in safety-sensitive environments.
    
    \item \textbf{Evidence-grounded explainability:} We provide span-level evidence attribution for each predicted code, enabling transparent and auditable decision support.
    
    \item \textbf{Agentic integration:} We illustrate how Symphony functions as a modular component within larger multi-agent systems, supporting long-running and autonomous clinical workflows.

    \item \textbf{Ontology scalability:} We demonstrate rapid onboarding of new coding systems and seamless adaptation to updates without retraining, enabling continuous alignment with evolving medical standards.
\end{enumerate}

Symphony for Medical Coding is accessible via a production-grade API and can be deployed either as a standalone coding service or as part of a broader agentic framework\footnote{Find more information on \href{https://docs.corti.ai/coding}{docs.corti.ai/coding} or try a demo on \href{https://console.corti.app}{console.corti.app}}.

\section{Background}

Clinical care generates vast amounts of unstructured data, including progress notes, lab reports, prescriptions, referral letters, treatment plans, and administrative documentation~\citep{liuNoteBloatImpacts2022}. Hidden within this unstructured data is information essential for medical decision-making, downstream operations, and medical science. To use this information, we must structure the data.

Healthcare systems structure the data by using medical coding systems: the process of translating clinical information into standardized, machine-readable codes that represent diagnoses, procedures, phenotypes, medications, and laboratory results. In the US, classification systems such as ICD-10, ICD-10-PCS, and CPT form the foundation of healthcare data systems, enabling reliable storage, automated validation, consistency checks, and auditing~\citep{bartaICD10CMOfficialCoding2009}. These taxonomies are developed and maintained by large expert committees and reflect broad clinical consensus. They are continuously updated to reflect new medical knowledge, changing practices, and evolving regulations, making them living references of clinical reality.

However, assigning these codes remains largely manual. It is both slow and expensive. In Scotland, a medical coder processes roughly 60 cases per day at 7–8 minutes per case~\citep{dongAutomatedClinicalCoding2022}. In the US, time per case ranges from under a minute for outpatient encounters to over thirty minutes for inpatient stays~\citep{tsengAdministrativeCostsAssociated2018, stanfillPreparingICD10CMPCS2014}. These bottlenecks create backlogs that can delay coding by months, sometimes even a full year~\citep{alonsoProblemsBarriersProcess2020}. The financial cost is equally stark. In 2012, billing and insurance-related activities (of which coding is a major component) cost the U.S. healthcare system an estimated \$471 billion (\$330B–\$597B)~\citep{jiwaniBillingInsurancerelatedAdministrative2014}. In systems where physicians perform coding themselves, this burden directly reduces time available for patient care.

Beyond being slow and costly, manual coding is error-prone. A systematic review of thirty-two studies found that the median accuracy of the primary diagnosis code was just 80.3\% (IQR: 63.3–94.1\%)~\citep{burnsSystematicReviewDischarge2012}. These errors have cascading consequences. Healthcare providers are under-reimbursed by governments and insurers. Clinical research that relies on coded data is contaminated with noise~\citep{jorgensenTimeorderedComorbidityCorrelations2021}. Patients may be directly harmed by an incorrect medical history; a pilot erroneously coded with depression, for example, may be denied licensure~\citep{dongAutomatedClinicalCoding2022}.

An AI system capable of reliably mapping free-text clinical narratives to standardized medical codes could address these challenges simultaneously. By converting unstructured documentation into validated, ontology-aligned representations, such a system would directly strengthen downstream analytics, quality reporting, and regulatory compliance~\citep{dongAutomatedClinicalCoding2022}. Perhaps most immediately, it would alleviate pressure on clinical documentation integrity (CDI) and revenue cycle management (RCM): two of the most labor-intensive and resource-constrained processes in modern healthcare operations.

Efforts to automate coding have progressed substantially, from early rule-based systems~\citep{farkasAutomaticConstructionRulebased2008, campbellComputerassistedClinicalCoding2020} through supervised machine learning approaches~\citep{mullenbachExplainablePredictionMedical2018, edinAutomatedMedicalCoding2023a} to more recent work with large language models~\citep{soroushLargeLanguageModels2024, kwanLargeLanguageModels2024}. Despite this progress, most existing approaches remain fundamentally constrained in two ways. First, they are locked to the coding system they were trained on. Because they learn statistical associations between clinical text and a fixed set of codes, they cannot generalize to other ontologies or accommodate yearly updates to classification systems without being retrained on newly labelled data. Second, they are opaque. The reasoning behind a suggested code is distributed across billions of learned parameters, offering clinicians no interpretable justification they can verify or challenge~\citep{edinUnsupervisedApproachAchieve2024a}. In combination, these limitations create a gap between benchmark performance and practical usability.

Beyond these methodological constraints, most prior work is limited in the scope of its evaluation. Studies typically report results on a single dataset drawn from one hospital and one clinical specialty, leaving it unclear whether performance generalises across settings~\citep{mullenbachExplainablePredictionMedical2018, dongAutomatedClinicalCoding2022, edinMedicalCodingLanguage}. This is a particularly acute concern for ICD-10 coding, where inpatient and outpatient encounters follow different coding guidelines, meaning that a system validated in one setting cannot be assumed to transfer to the other~\citep{bartaICD10CMOfficialCoding2009}. The field's heavy reliance on MIMIC, a publicly available inpatient dataset, has further narrowed the evidence base, raising the risk that reported results reflect overfitting to the idiosyncrasies of a single institution rather than genuine coding ability ~\citep{johnsonMIMICIIIFreelyAccessible2016, johnsonMIMICIVFreelyAccessible2023}.

We propose Symphony for Medical Coding, a next-generation system that can adapt to arbitrary coding ontologies without retraining and that provides span-level evidence attribution for every predicted code, making its reasoning transparent and auditable. We evaluate Symphony across five datasets spanning both inpatient and outpatient settings, multiple specialties, and two countries (the United States and the United Kingdom), providing the most comprehensive assessment of an automated coding system to date.

\section{Methods}
\subsection{Symphony for Medical Coding}

Symphony for medical coding is an extension to the CLH framework~\citep{motzfeldtCodeHumansMultiAgent2025}. The CLH framework formulates medical coding as a structured, multi-step reasoning process rather than a single-step classification task. Inspired by the workflow of expert human coders, CLH assigns codes across multiple subsequent steps, explicitly incorporating official coding resources and conventions.

The framework consists of four pillars:
\begin{enumerate}
    \item \textbf{Evidence Extraction:} Identify which medical codes should be coded within a clinical text.
    \item \textbf{Index Navigation:} For each medical code, find the term in the alphabetic index that best describes it. This term will refer to a location within the code hierarchy.
    \item \textbf{Tabular Validation:} From the location, navigate in the hierarchy to find the most precise code.
    \item \textbf{Code Reconciliation:} In the end, consider all the identified codes and remove those that are mutually exclusive.
\end{enumerate}

Each stage is implemented as an LLM-powered agent with access to structured coding resources, enabling ontology-aware reasoning over the full medical coding label space (approximately 70,000 codes for ICD-10-CM). By explicitly modeling the medical coding steps followed by human experts, CLH supports open-set coding and improves robustness on rare and long-tail codes compared to conventional supervised multi-label classifiers.

Symphony introduces several improvements over the original CLH implementation. These include more a priori knowledge, and stronger reasoning and search capabilities. These improvements result in substantial performance gains.

\begin{figure}[htbp]
  \centering
  \includegraphics[width=\linewidth]{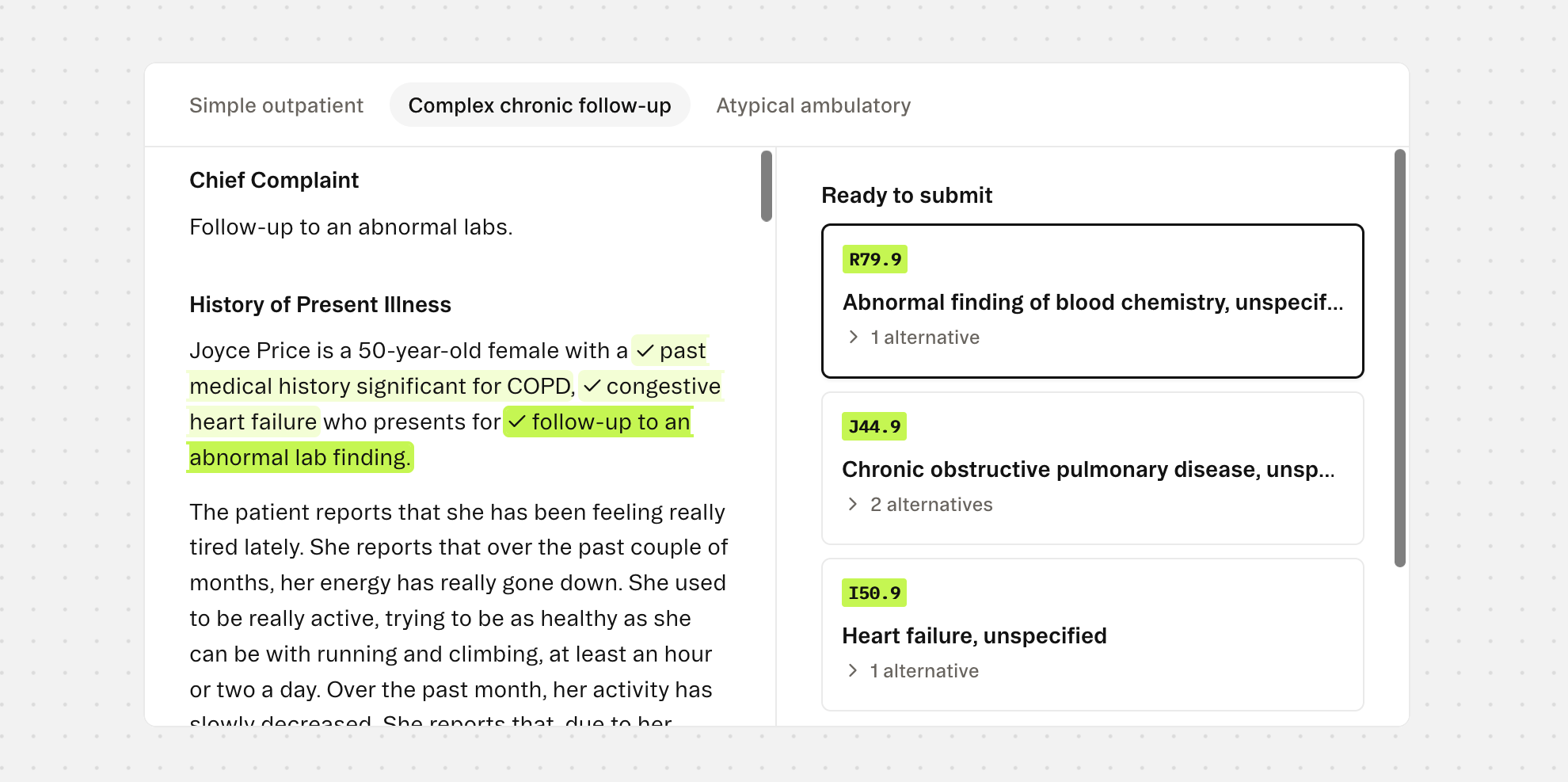}
  \caption{A sample user interface showing the evidence spans provided by Symphony for Medical Coding.}
\label{fig:explainable_coding}
\end{figure}

Since Symphony extracts the mentions of diagnoses and procedures that should be coded, it naturally provides span-level evidence for each code. This explainability makes it particularly well-suited for integration into larger clinical coding pipelines (see Fig. \ref{fig:explainable_coding}): enabling transparent validation, auditing, and human-in-the-loop review, which are critical requirements in safety-sensitive and regulatory environments. This explicit grounding enables downstream systems and medical coders to efficiently verify decisions, resolve ambiguities, and ensure compliance with coding guidelines. As a result, Symphony can function not only as an autonomous coding system but also as a reliable decision-support component within hybrid workflows that combine automation with expert oversight.

Because Symphony breaks medical coding into explicit steps — extracting evidence, finding candidate codes, and verifying guidelines — its intermediate outputs are structured and well-defined. This makes it a natural building block in larger multi-agent clinical systems\footnote{Symphony for Medical Coding is available, not only as an API endpoint, but also as MCPs that can plug into a multi-agent system (\href{https://www.corti.ai/agents}{See Corti's Agents Library}).}, where upstream agents can feed it structured inputs and downstream agents can act directly on its outputs, without duplicating reasoning or re-processing raw clinical documentation.

\subsection{Data}
We evaluate Symphony on five datasets that collectively span inpatient and outpatient settings, multiple clinical specialties, two countries, and both public and proprietary sources.

\begin{table}[ht]
\centering
\label{tab:datasets}
\begin{tabular}{llrr}
\toprule
\textbf{Dataset} & \textbf{Source} & \textbf{\text{\#}Diagnosis Codes} & \textbf{\text{\#}Procedure Codes} \\
\midrule
ACI          & Academia   & 225  &  - \\
MDACE             & Academia   & 904  & 118 \\
Neurology (NEURO) & Private  & 667  & -  \\
Emergency Departement (ED)         & Private    & 11,148 & 655 \\
Ambulatory (AMB) & Private    & 17,561 & 2,640 \\
\bottomrule
\\
\end{tabular}
\caption{Summary of evaluation datasets. The number of unique diagnosis and procedure codes reflects the target code space per dataset across all splits. Public datasets come from prior academic benchmarks; proprietary datasets are drawn from production clinical workflows.}
\end{table}

\subsubsection{Public datasets}
\paragraph{ACI}
ACI-BENCH is a small outpatient dataset originally developed for evaluating AI scribes~\citep{yimAcibenchNovelAmbient2023}. Each case consists of a transcribed simulated dialogue between a physician and patient, from which a separate physician wrote a clinical note. \citet{yuanReliableClinicalCoding2025} subsequently engaged professional medical coders to assign ICD-10-CM codes to these notes. We used the clinical note and the ICD-10-CM codes for the evaluation. While limited in scale, ACI-BENCH provides a controlled outpatient evaluation setting with high-quality annotations.

\paragraph{MDACE}
MDACE comprises 302 intensive care unit encounters drawn from MIMIC-III~\citep{chengMDACEMIMICDocuments2023}. Professional medical coders reannotated each encounter with ICD-10-CM and ICD-10-PCS codes, and annotations were cross-validated across coders to ensure quality. Each encounter contains multiple clinical note types. Previous work has evaluated on individual notes~\citep{motzfeldtCodeHumansMultiAgent2025,yuanReliableClinicalCoding2025}, but we concatenate all notes within each encounter into a single document, as we found that codes were frequently missing when individual note types were evaluated in isolation.

\subsubsection{Proprietary datasets}

To evaluate performance in real-world clinical settings, we include two proprietary datasets reflecting distinct care environments.

\paragraph{Emergency department (ED)}
The first dataset comprises 563,153 clinical notes containing 11,148 unique ICD-10-CM codes from an emergency department at a large U.S. provider system. The ED setting presents a particularly demanding test for automated coding. Documentation is often fragmented and symptom-driven, produced under time pressure with high diagnostic uncertainty. Coding decisions frequently depend on incomplete or provisional information, which makes coding particularly challenging.

\paragraph{Ambulatory (AMB)}
The second dataset consists of 2,250,380 ambulatory (outpatient) clinical notes containing 17,561 unique ICD-10-CM codes from the same U.S. provider system. In contrast to the ED setting, ambulatory care reflects scheduled, longitudinal interactions with more structured documentation focused on chronic disease management and preventive care. Coding patterns tend to be denser and more stable, often involving broad sets of comorbidities. This dataset tests the system's ability to handle fine-grained diagnostic distinctions across a large code space.

\paragraph{Neurology (NEURO)}
The third proprietary dataset comprises 10,675 clinical notes containing 667 unique codes from a neurology department within the UK National Health Service. This dataset uses the NHS version of ICD-10, which differs from the ICD-10-CM system used in the U.S. datasets in both structure and code granularity. We use it specifically to demonstrate Symphony's ability to adapt to an entirely different coding ontology without retraining. It also introduces a subspecialty setting with a concentrated, technically demanding code distribution. Finally, it tests cross-national generalisability, as UK clinical documentation follows different conventions and terminology than its U.S. counterparts.

\section{Experiments}

\subsection{Evaluation setting}

Prior work in medical coding often simplifies the task by restricting prediction to a predefined subset of codes \citep{edinAutomatedMedicalCoding2023a}. In practice, this reduces the output space from the full classification system to a much smaller label set, often on the order of only 1,000 codes. While this setup enables the use of standard supervised classification approaches, it substantially alters the nature of the problem.

This restriction introduces two key limitations. First, it artificially makes the problem easier by excluding neighboring codes from the evaluation. For example, there are numerous medical codes for \textit{diabetes mellitus}, each specifying the type and any complications. If only one or two of these appear in the filtered test set, the system is never penalised for confusing closely related codes that it would encounter in practice. Reported accuracy, therefore, overstates what could be expected in deployment against the full ontology. Second, it assumes prior knowledge of the relevant label space, which is not always available in deployment settings where the system must operate over the full, evolving classification system. As a result, performance reported under restricted label settings may significantly overestimate real-world utility.

Symphony for Medical Coding can easily adapt to a sub-selection of codes. We evaluate systems under two complementary settings:

\begin{itemize}
    \item \textbf{Restricted code system evaluation:} Models predict over a predefined subset of codes, following common practice in prior work. This setting enables direct comparison to existing benchmarks.
    
    \item \textbf{Full code system evaluation:} Models predict over the complete coding system without label restrictions, reflecting real-world deployment requirements.
\end{itemize}

We evaluate each system five times and report its mean and standard deviation. Some of the results we do not reproduce, but instead copy from their respective papers. Most of these studies did not evaluate their models multiple times; therefore, we cannot report the mean and standard deviation.

\subsubsection{Evaluation metrics}
Similar to prior work, we evaluate all systems using precision, recall, and F1 score~\citep{edinAutomatedMedicalCoding2023a,zhangGeneralKnowledgeInjection2025}.

Precision measures the proportion of predicted codes that are correct, i.e., how much you can trust a prediction:
\begin{equation}
    \text{Precision (P)} = \frac{\text{TP}}{\text{TP} + \text{FP}}
\end{equation}
where TP denotes true positives, and FP denotes false positives. A system with low precision frequently assigns codes that do not belong to the encounter, generating noise for downstream billing and analytics.

Recall measures the proportion of true codes that are successfully identified:
\begin{equation}
    \text{Recall (R)} = \frac{\text{TP}}{\text{TP} + \text{FN}}
\end{equation}
where FN denotes false negatives. A system with low recall misses codes that should have been assigned, leading to incomplete clinical records and potential under-reimbursement.

The F1 score combines both into a single measure via their harmonic mean:
\begin{equation}
    \text{F1} = 2 \cdot \frac{\text{Precision} \cdot \text{Recall}}{\text{Precision} + \text{Recall}}
\end{equation}
The harmonic mean ensures that a system must perform well on both precision and recall to achieve a high F1; strong performance on one cannot compensate for poor performance on the other.

These metrics can be aggregated across codes in two ways. Micro-averaging pools all true positives, false positives, and false negatives across codes before computing a single score, effectively weighting each prediction equally. Macro-averaging instead computes the metric independently for each code and then takes the unweighted mean, giving equal weight to every code regardless of its frequency. Both are widely reported in the medical coding literature, and the choice between them can substantially affect how system performance is perceived.

We report micro-averaged scores. This choice is motivated by two properties of our evaluation setting. First, macro-averaging has low statistical power when most classes are rare and test sets are small, because a single example being correctly or incorrectly classified can swing a class-level F1 from 0 to 1, or vice versa. Second, macro-averaging can mask systematic overprediction. If a system predicts I10 (essential hypertension) for every encounter, the precision for that code collapses, but because I10 is only one class out of thousands, its effect on the macro-average is negligible. Micro-averaging, by contrast, counts every false positive equally, making such behaviour immediately visible in the overall score.

These metrics should be interpreted with some caution, as the target labels themselves are not always reliable. Medical coding is a subjective, guideline-driven process, and prior studies have shown that even professional coders often disagree on the correct codes \citep{chengMDACEMIMICDocuments2023}. This implies that the reference annotations used for evaluation are inherently uncertain and may not reflect a single, unambiguous ground truth. As a result, there exists a practical upper bound on achievable F1 scores when evaluating against a single set of annotations, since even expert coders would not consistently agree with that reference. Model performance should therefore be interpreted in the context of this label noise, particularly when disagreements arise from alternative yet clinically plausible coding decisions.

Furthermore, the long-tailed distribution of medical codes complicates the interpretation of aggregate metrics such as F1. A small number of common codes account for a large fraction of encounters, while the majority of codes appear only rarely. In practice, this means that a system can achieve near-perfect accuracy on high-volume, well-defined coding patterns while still receiving a moderate overall F1 score due to errors on rare or highly specific codes. Consequently, F1 alone does not fully capture the extent to which a system can automate real-world coding workloads. High performance on frequent and clinically critical code combinations may enable substantial automation in practice, even if the aggregate F1 score suggests room for improvement.

\begin{figure}[h]
  \centering
  \includegraphics[width=\linewidth]{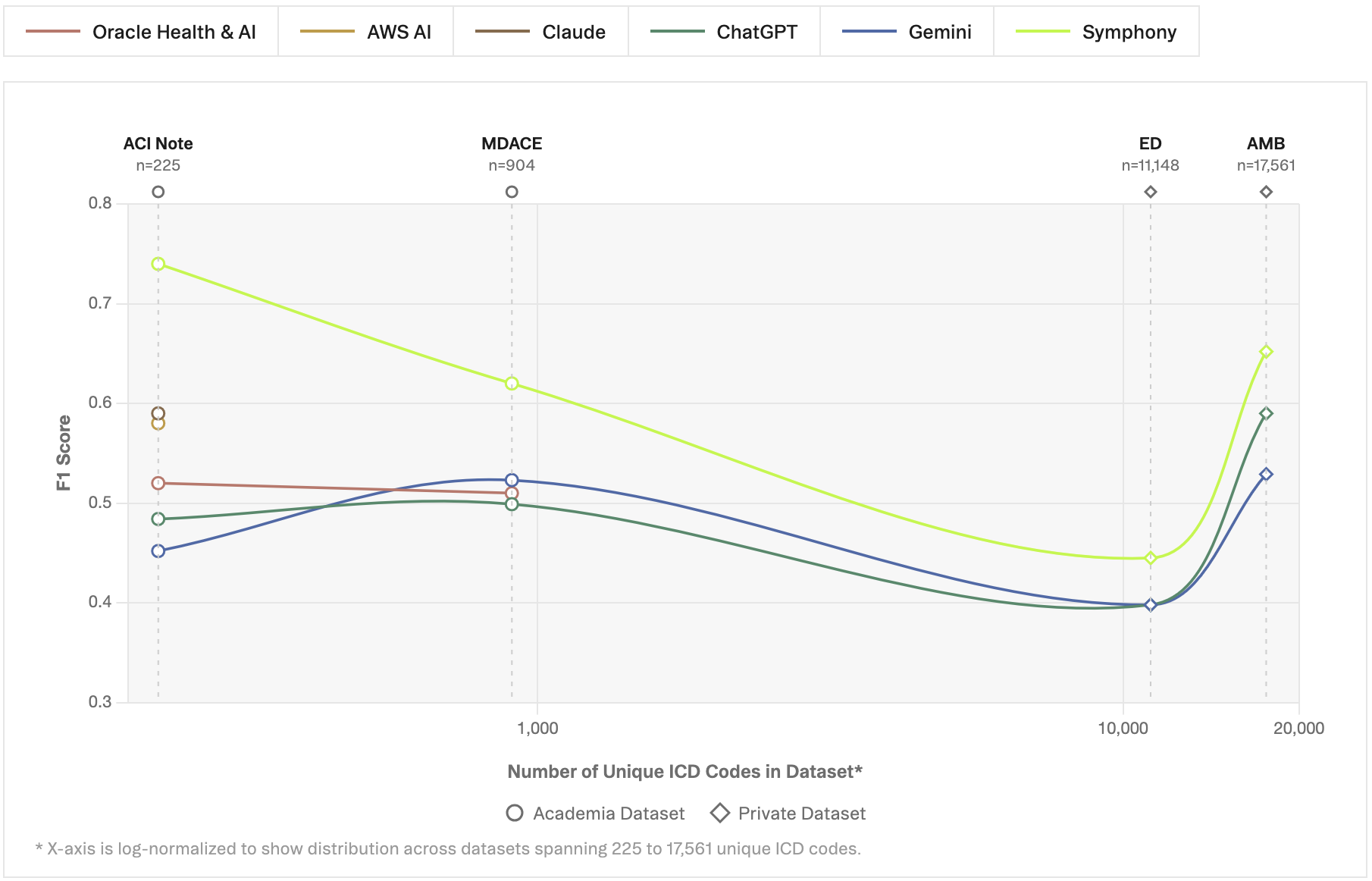}
  \caption{F1 performance as a function of code space size (log-scaled) from \citet{yuanReliableClinicalCoding2025a, zhengMedDCRLearningDesign2025} and experiments in this paper. Each point corresponds to a dataset with a distinct number of unique codes. Symphony demonstrates superior performance across label space complexity.}
\label{fig:results}
\end{figure}

\subsection{Results}

\subsubsection{Restricted code system evaluation}

Table~\ref{tab:icd-10-cm_micro_scores1} presents results under the restricted code system setting. Across both datasets, Symphony achieves the highest overall performance, significantly outperforming prior approaches.

We group the compared methods into three categories based on their underlying approach. \textbf{Fine-tuned models} are supervised approaches trained directly for code prediction over a fixed label set. Claude was fine-tuned on ACI, while PLM-CA and PLM-ICD were on MIMIC-IV~\citep{nguyenMimicIVICDNewBenchmark2023}. \textbf{Agent-based methods} use large language models with prompting strategies such as chain-of-thought and self-consistency to perform coding through reasoning, without explicit task-specific training. \textbf{Workflow-based methods} extend this paradigm by decomposing the coding task into multiple coordinated steps or agents, often combining reasoning, retrieval, and verification in structured pipelines.

On the ACI dataset, Symphony reaches an F1 score of 0.74, improving substantially over the strongest baseline, MedDCR (0.52), developed by Oracle Health \& AI on the OpenAI GPT family of models, and over the best fine-tuned model, Claude (0.58), developed by the AWS AI team on the Anthropic Claude model family. This gain is driven by a strong balance between recall (0.81) and precision (0.68), indicating that Symphony can both identify relevant codes and maintain high precision in its predictions.

On MDACE, Symphony similarly achieves the best performance with an F1 score of 0.58, surpassing MedDCR (0.51), developed by Oracle Health \& AI on GPT-based models, and fine-tuned baselines such as PLM-ICD (0.48). Notably, Symphony maintains a more balanced precision-recall trade-off compared to prior methods, which often exhibit either high recall with low precision or vice versa.

Across both datasets, prior agentic and workflow-based approaches tend to suffer from low precision, despite achieving moderate recall. In contrast, Symphony consistently improves precision while maintaining competitive or higher recall, resulting in substantially higher F1 scores.

Finally, the low standard deviation across runs indicates that Symphony produces stable and reproducible results.
\begin{table}[h]
\centering
\caption{\textbf{Restricted code system evaluation}. Symphony in comparison to prior approaches, including fine-tuned models~\citep{huangPLMICDAutomaticICD2022, edinUnsupervisedApproachAchieve2024a, yuanReliableClinicalCoding2025a} (with Claude based on the Anthropic Claude model family developed by AWS AI), agent-based methods, and workflow-based systems such as MedDCR~\citep{zhengMedDCRLearningDesign2025} (built by Oracle Health \& AI on the OpenAI GPT model family) and related approaches~\citep{kwanLargeLanguageModels2024, liExploringLLMMultiAgents2024}. We evaluated Symphony five times. The table shows the mean value and the standard deviation across runs. The results for the other models are taken from their respective papers. The best F1-score is shown in bold.}\vspace{0.5cm}
\label{tab:icd-10-cm_micro_scores1}
\resizebox{\textwidth}{!}{
\begin{tabular}{llcccccc}
\toprule
 & & \multicolumn{3}{c}{\textbf{ACI}} & \multicolumn{3}{c}{\textbf{MDACE}}  \\
\cmidrule(lr){3-5} \cmidrule(lr){6-8} 
\textbf{Method} & \textbf{Model} & \textbf{R} & \textbf{P} & \textbf{F1} & \textbf{R} & \textbf{P} & \textbf{F1} \\
\midrule
 \multirow{3}{*}{Fine-tune} & PLM-ICD & 0.41 & 0.43 & 0.42 & 0.47 & 0.49 & 0.48 \\
 & PLM-CA & 0.42 & 0.44 & 0.43 & 0.45 & 0.46 & 0.45 \\
 & Claude & - & - & 0.58 & - & - & - \\
\midrule
 \multirow{5}{*}{Agent} & CoT & 0.50 & 0.35 & 0.41 & 0.31 & 0.30 & 0.30 \\
 & CoT-SC & 0.59 & 0.36 & 0.44 & 0.43 & 0.39 & 0.41 \\
 & MulDe & 0.65 & 0.16 & 0.25 & 0.40 & 0.21 & 0.28 \\
 & Judge & 0.64 & 0.22 & 0.33 & 0.53 & 0.26 & 0.35 \\
 & ADAS & 0.59 & 0.28 & 0.43 & 0.51 & 0.37 & 0.43 \\
\midrule
 \multirow{4}{*}{Workflow} & RRS & 0.52 & 0.26 & 0.35 & 0.30 & 0.24 & 0.27 \\
 & MAC & 0.50 & 0.23 & 0.31 & 0.31 & 0.27 & 0.29 \\
 & MedDCR & 0.67 & 0.43 & 0.52 & 0.59 & 0.46 & 0.51 \\
 
 \\
 & \textbf{Symphony} & $0.81${\scriptsize $\pm 0.01$} & $0.68${\scriptsize $\pm 0.01$} & $\mathbf{0.74}${\scriptsize $\boldsymbol{\pm 0.01}$} & $0.66${\scriptsize $\pm 0.02$} & $0.59${\scriptsize $\pm 0.02$} & $\mathbf{0.62}${\scriptsize $\boldsymbol{\pm 0.02}$}\\
\bottomrule
\end{tabular}
}
\end{table}
\subsubsection{Full code system evaluation}

Table~\ref{tab:icd-10-cm_micro_scores2_part1} and \ref{tab:icd-10-cm_micro_scores2_part2} present results under the full code system setting, where models are required to predict over the complete coding space without label restrictions. To the best of our knowledge, such evaluations have not been reported by model providers, despite both Anthropic and OpenAI positioning their models as ready for healthcare applications.

We compare three classes of systems\footnote{Despite multiple attempts to obtain a HIPAA-compliant deployment option from Anthropic, we were unable to process PHI-containing datasets with Claude.}: foundation models without tools, foundation models augmented with tools, and specialized coding systems. The \textit{with tools} setting for Claude leverages the Anthropic Model Context Protocol (MCP) adapted for medical coding, enabling structured interaction with external coding resources. While tool augmentation improves performance in some cases, results remain inconsistent across datasets and models.

Across all datasets, Symphony achieves the strongest and most consistent performance, outperforming both standalone foundation models and tool-augmented variants. In contrast, general-purpose models tend to exhibit high recall but substantially lower precision, particularly in the full-classification system setting, reflecting the challenge of operating over large and highly imbalanced label spaces. These results highlight the gap between general-purpose LLM capabilities and the requirements of real-world medical coding, and demonstrate the importance of structured, ontology-aware systems for scalable deployment.

\begin{table}[h]
\centering
\caption{\textbf{Full code system evaluation} on public datasets. F1-scores (multiplied by 100) across models: Anthropic Claude (Opus 4.6), OpenAI GPT (5.4), Google Gemini (3.1 Pro), and Corti Symphony. We evaluated each model five times. The table shows the mean value and the standard deviation across runs. The best score for each metric is shown in bold. $^\dagger$Despite multiple attempts over an extended period, we were unable to obtain a HIPAA-compliant deployment from Anthropic, restricting evaluation of Claude to the non-PHI ACI dataset.}\vspace{0.5cm}
\label{tab:icd-10-cm_micro_scores2_part1}
\resizebox{\textwidth}{!}{
\begin{tabular}{llcccccc}
\toprule
 & & \multicolumn{3}{c}{\textbf{ACI}} & \multicolumn{3}{c}{\textbf{MDACE}} \\
\cmidrule(lr){3-5} \cmidrule(lr){6-8}
\textbf{Method} & \textbf{Model} & \textbf{R} & \textbf{P} & \textbf{F1} & \textbf{R} & \textbf{P} & \textbf{F1} \\
\midrule
\multirow{2}{*}{No tools} & Claude$^\dagger$ & 51.0{\scriptsize $\pm 4.1$} & 45.3{\scriptsize $\pm 8.8$} & 47.6{\scriptsize $\pm 5.2$} & - & - & - \\
& ChatGPT & 74.2{\scriptsize $\pm 1.8$} & 42.5{\scriptsize $\pm 1.2$} & 54.0{\scriptsize $\pm 1.4$} & 66.0{\scriptsize $\pm 1.9$} & 43.1{\scriptsize $\pm 1.3$} & 52.1{\scriptsize $\pm 1.4$} \\
\midrule
\multirow{2}{*}{With tools} & Claude$^\dagger$ & 60.3{\scriptsize $\pm 8.47$} & 58.13{\scriptsize $\pm 8.47$} & 59.0{\scriptsize $\pm 6.1$} & - & - & - \\
& ChatGPT & 72.4{\scriptsize $\pm 0.9$} & 36.4{\scriptsize $\pm 0.9$} & 48.4{\scriptsize $\pm 0.9$} & 56.0{\scriptsize $\pm 1.5$} & 45.1{\scriptsize $\pm 1.3$} & 49.9{\scriptsize $\pm 1.3$} \\

& Gemini & 76.5{\scriptsize $\pm 1.1$} & 32.1{\scriptsize $\pm 0.1$} & 45.2{\scriptsize $\pm 0.1$} & 60.0{\scriptsize $\pm 0.6$} & 46.4{\scriptsize$\pm 0.4$} & 52.3{\scriptsize $\pm 0.5$} \\
\\
\multirow{2}{*}{Corti} & CLH & 56.0{\scriptsize $\pm 1.0$} & 58.5{\scriptsize $\pm 1.3$} & 57.2{\scriptsize $\pm 1.0$} & 30.7{\scriptsize $\pm 0.7$} & 51.9{\scriptsize $\pm 0.8$} & 38.5{\scriptsize $\pm 0.5$} \\
& \textbf{Symphony} & 75.0{\scriptsize $\pm 1.2$} & 56.0{\scriptsize $\pm 2.0$} & \textbf{64.1}{\scriptsize $\boldsymbol{\pm 1.7}$} & 61.8{\scriptsize $\pm 1.7$} & 49.6{\scriptsize $\pm 1.0$} & \textbf{55.0}{\scriptsize $\boldsymbol{\pm 0.2}$} \\
\bottomrule
\end{tabular}
}
\end{table}

\begin{table}[h]
\centering
\caption{\textbf{Full code system evaluation} on Corti's proprietary datasets. F1-scores (multiplied by 100) across models: Anthropic Claude (Opus 4.6), OpenAI GPT (5.4), Google Gemini (3.1 Pro), and Corti Symphony. We evaluated each model five times. The table shows the mean value and the standard deviation across runs. The best score for each metric is shown in bold. $^\dagger$Despite multiple attempts over an extended period, we were unable to obtain a HIPAA-compliant deployment from Anthropic, restricting evaluation of Claude to the non-PHI ACI dataset.}\vspace{0.5cm}
\label{tab:icd-10-cm_micro_scores2_part2}
\resizebox{\textwidth}{!}{
\begin{tabular}{llcccccc}
\toprule
 & & \multicolumn{3}{c}{\textbf{AMB}} & \multicolumn{3}{c}{\textbf{ED}} \\
\cmidrule(lr){3-5} \cmidrule(lr){6-8}
\textbf{Method} & \textbf{Model} & \textbf{R} & \textbf{P} & \textbf{F1} & \textbf{R} & \textbf{P} & \textbf{F1} \\
\midrule
\multirow{2}{*}{No tools} & Claude$^\dagger$ & - & - & - & - & - & - \\
& ChatGPT & 77.0{\scriptsize $\pm 0.4$} & 47.6{\scriptsize $\pm 0.8$} & 58.8{\scriptsize $\pm 0.6$} & 42.3{\scriptsize $\pm 0.7$} & 42.8{\scriptsize $\pm 0.5$} & 42.5{\scriptsize $\pm 0.5$} \\
\midrule
\multirow{2}{*}{With tools} & Claude$^\dagger$ & - & - & - & - & - & - \\
& ChatGPT & 70.4{\scriptsize $\pm 0.6$} & 50.8{\scriptsize $\pm 0.4$} & 59.0{\scriptsize $\pm 0.4$} & 46.6{\scriptsize $\pm 0.3$} & 34.7{\scriptsize $\pm 0.4$} & 39.8{\scriptsize $\pm 0.3$} \\
& Gemini & 42.6{\scriptsize $\pm 0.2$} & 67.8{\scriptsize $\pm 0.7$} & 52.3{\scriptsize $\pm 0.5$} & 27.1{\scriptsize $\pm 0.3$} & 35.8{\scriptsize $\pm 0.4$} & 39.8{\scriptsize $\pm 0.3$} \\
\\
\multirow{2}{*}{Corti} & CLH & 51.1{\scriptsize $\pm 0.6$} & 62.0{\scriptsize $\pm 0.2$} & 56.0{\scriptsize $\pm 0.4$} & 26.5{\scriptsize $\pm 0.5$} & 49.9{\scriptsize $\pm 0.8$} & 34.6{\scriptsize $\pm 0.6$} \\
& \textbf{Symphony} & 64.4{\scriptsize $\pm 0.3$} & 66.0{\scriptsize $\pm 0.5$} & \textbf{65.2}{\scriptsize $\boldsymbol{\pm 0.2}$} & 37.6{\scriptsize $\pm 0.3$} & 54.3{\scriptsize $\pm 0.8$} & \textbf{44.5}{\scriptsize $\boldsymbol{\pm 0.5}$} \\
\bottomrule
\end{tabular}
}
\end{table}

In Table \ref{tab:icd-10-uk_micro_scores} we evaluated Symphony on the U.K. NEURO coding dataset. We compare Symphony against GPT-based models, as these were the only foundation models we could evaluate in a GDPR-compliant setting. Given the similar performance patterns observed across ChatGPT, Claude, and Gemini in the experiments above, we consider ChatGPT to be a representative baseline for this setting.

These results highlight Symphony’s ability to generalize across coding systems and geographic standards. Despite differences in coding guidelines, terminology, and clinical documentation practices between U.S. and U.K. settings, Symphony maintains strong performance without retraining, achieving a substantial improvement over GPT (32.9 vs. 18.1 F1). This suggests that Symphony’s ontology-aware and reasoning-driven approach is not tied to a specific coding system, but instead captures underlying coding principles that transfer effectively across domains. The performance gap further indicates that structured workflows are particularly important when adapting to new coding regimes, where reliance on parametric knowledge alone is insufficient.

A further insight from the U.K. evaluation is the presence of systematic under-specification in the reference annotations. In additional analyses, we observe that a substantial portion of disagreements arises from differences in specificity rather than incorrect clinical interpretation. In many cases, the reference data assigns broader, unspecified codes where the clinical documentation supports more precise alternatives selected by Symphony. This pattern, commonly observed in NHS coding practice, reflects a tendency toward conservative coding choices rather than maximal specificity. As a result, part of the residual error measured at the code level may stem from limitations in the reference annotations rather than model performance. This further highlights the value of structured, reasoning-based systems in promoting consistent and clinically precise coding.

\begin{table}
\centering
\caption{\textbf{Full code system evaluation for ICD-10 (UK)}. F1-scores (multiplied by 100) on OpenAI GPT (5.4) and Corti Symphony.
We evaluated each model five times. The table shows the mean value and the standard deviation
across runs. The best score for each metric is shown in bold. The table shows the mean value and the standard deviation across runs. The best score for each metric is shown in bold.}\vspace{0.5cm}
\label{tab:icd-10-uk_micro_scores}
\begin{tabular}{lccc}
\toprule
 & \multicolumn{3}{c}{\textbf{NEURO}} \\
\cmidrule(lr){2-4}
\textbf{Model} & \textbf{R} & \textbf{P} & \textbf{F1} \\
\midrule
ChatGPT & 21.3{\scriptsize $\pm 0.3$} & 15.8{\scriptsize $\pm 0.2$} & 18.1{\scriptsize $\pm 0.2$} \\
\\
\textbf{Symphony} & $49.6${\scriptsize $\pm 0.4$} & $24.6${\scriptsize $\pm 0.5$} & $\mathbf{32.9}${\scriptsize $\boldsymbol{\pm 0.5}$} \\
\bottomrule
\end{tabular}
\end{table}

\begin{table}[h!]
\centering
\caption{\textbf{Full procedure coding system evaluation}. F1-scores (multiplied by 100) on OpenAI GPT (5.4) and Corti Symphony. We evaluated each model five times. The table shows the mean value and the standard deviation across runs. The best score for each metric is shown in bold. $^\dagger$Symphony for CPT is in \texttt{beta}. The results may not generalize well to some specialties (updates are coming soon).}\vspace{0.5cm}
\label{tab:icd-10-pcs_micro_scores}
\resizebox{\textwidth}{!}{
\begin{tabular}{lccccccccc}
\toprule
& \multicolumn{3}{c}{\textbf{ICD-10-PCS}} & \multicolumn{6}{c}{\textbf{CPT$^{\dagger}$}} \\
\cmidrule(lr){2-4} \cmidrule(lr){5-10}
 & \multicolumn{3}{c}{\textbf{MDACE}} & \multicolumn{3}{c}{\textbf{AMB}} & \multicolumn{3}{c}{\textbf{ED}}\\
\cmidrule(lr){2-4} \cmidrule(lr){5-7} \cmidrule(lr){8-10}
\textbf{Model} & \textbf{R} & \textbf{P} & \textbf{F1} & \textbf{R} & \textbf{P} & \textbf{F1}  & \textbf{R} & \textbf{P} & \textbf{F1}\\
\midrule

ChatGPT 
& 34.8{\scriptsize $\pm 2.4$} & 26.4{\scriptsize $\pm 3.1$} & 30.0{\scriptsize $\pm 2.9$} 
& 31.1{\scriptsize $\pm 0.6$} & 66.0{\scriptsize $\pm 1.8$}  & 42.2{\scriptsize $\pm 0.6$}& 21.0{\scriptsize $\pm 1.1$} & 20.5{\scriptsize $\pm 0.8$} & 20.8{\scriptsize $\pm 0.9$}\\

\\
\textbf{Symphony} 
& $37.5${\scriptsize $\pm 0.8$} & $37.0${\scriptsize $\pm 0.6$} & $\mathbf{37.3}${\scriptsize $\boldsymbol{\pm 0.4}$} 
& 51.8{\scriptsize $\pm 0.4$} & 50.8{\scriptsize $\pm 0.4$} & $\mathbf{51.3}${\scriptsize $\boldsymbol{\pm 0.4}$}  & 48.8{\scriptsize $\pm 0.6$} & 34.2{\scriptsize $\pm 0.3$} & $\mathbf{40.2}${\scriptsize $\boldsymbol{\pm 0.4}$} \\

\bottomrule
\end{tabular}
}
\end{table}

\subsubsection{Procedure coding evaluation}

While the previous results were focused on diagnosis coding, Table \ref{tab:icd-10-pcs_micro_scores} presents results on ICD-10-PCS coding using the MDACE dataset and CPT using the AMB and ED dataset. We compare Symphony against GPT-based models, as these were the only foundation models we could evaluate in a HIPAA-compliant setting for procedure coding tasks. Given the similar performance patterns observed across GPT, Claude, and Gemini in the experiments above, we consider GPT to be a representative baseline for this setting.

Symphony achieves the best overall performance across systems. On ICD-10-PCS (MDACE), Symphony attains an F1 score of 37.3, outperforming GPT (30.0). While GPT achieves moderate recall, it exhibits a pronounced imbalance with substantially lower precision, limiting overall performance. In contrast, Symphony maintains a near-equal balance between recall (37.5) and precision (37.0), resulting in a significantly higher F1 score. The operation notes are missing in MDACE, which could be the explanation for both scores being lower than for the other coding systems~\cite{chengMDACEMIMICDocuments2023}.

This advantage extends to CPT coding. On the ambulatory dataset, Symphony achieves an F1 score of 51.3 compared to 42.2 for GPT, while maintaining balanced recall (51.8) and precision (50.8). On the ED dataset, Symphony similarly outperforms GPT (40.2 vs. 20.8), again with a more balanced precision-recall profile. These results highlight both the increased difficulty of procedure coding compared to diagnosis coding and the importance of structured reasoning and ontology-aware workflows when operating over large and heterogeneous coding systems such as ICD-10-PCS and CPT.

\subsubsection{Evidence-span Explainability Analysis}
\begin{wraptable}{r}{0.45\linewidth}
\vspace{-0.5cm}
\centering
\caption{\textbf{Evidence span explainability on MDACE}. Key metrics evaluating alignment between predicted and human-annotated evidence spans.}
\label{tab:evidence_spans}
\begin{tabular}{lc}
\toprule
\textbf{Metric} & \textbf{Value} \\
\midrule
Coverage ($\geq 1$ predicted span) & $98.9\%$ \\
Hit rate (IoU > 0) & $73.7\%$ \\
Hit rate (IoU > 0.5) & $42.5\%$ \\
\midrule
Character-level F1 & $0.459$ \\
ROUGE-L F1 & $0.506$ \\
Span IoU (avg) & $0.471$ \\
\bottomrule
\end{tabular}
\vspace{-0.2cm}
\end{wraptable}
Each clinical note in MDACE is annotated with evidence spans: spans of text that contain the evidence for a code~\citep{chengMDACEMIMICDocuments2023}. We compare Symphony's extracted evidence spans with those annotated in the dataset. Table~\ref{tab:evidence_spans} summarizes the quality of the evidence spans produced by Symphony on the MDACE dataset. The system provides evidence for nearly all correct predictions, achieving a coverage of 98.9\%. Among these, 73.7\% of predicted spans overlap with human-annotated evidence, with 42.5\% showing substantial alignment (IoU > 0.5). At the token level, Symphony attains a character-level F1 of 0.459 and a ROUGE-L F1 of 0.506, indicating that the predicted evidence captures the relevant clinical content while allowing for some flexibility in span boundaries. The average span IoU of 0.471 further supports that the model frequently highlights similar regions of text as human annotators. Note, that the human-annotated evidence spans in MDACE do not cover all mentions of a code, and often do not contain full sentences~\citep{chengMDACEMIMICDocuments2023,beckhCanEnsemblesImprove2025,beckhAnatomyEvidenceInvestigation2025,douglasLessMoreExplainable2025}. The true scores of Symphony's extracted evidence spans is therefore higher than the scores indicate. Overall, these results show that Symphony produces reliable and well-grounded evidence, with most discrepancies arising from differences in annotation granularity rather than incorrect reasoning.

\section{Conclusion}

We introduced Symphony for Medical Coding, an ontology-aware agentic system for scalable, explainable, and deployment-ready medical coding. Across both restricted-label and full-classification system evaluations, Symphony achieves state-of-the-art performance on public benchmarks and on real-world provider data, while maintaining the high precision required for safety-sensitive clinical workflows. In contrast to prior supervised approaches that operate over fixed subsets of codes, Symphony supports reasoning over full and evolving coding systems without retraining, and provides span-level evidence attribution for transparent and auditable decision support.

These results show that medical coding benefits from structured, ontology-aware reasoning rather than closed-set prediction alone. They also demonstrate that Symphony can serve not only as a high-performance coding engine, but as a modular component within broader autonomous clinical systems.

\section*{Acknowledgements}
We thank Casper Christensen, Majed Sharif, Maxwell White, Nicklas Frahm, Maciej Tatarski, Henrik Cullen, and Dan Engel for their valuable contributions, insightful discussions, and support throughout the development and evaluation of this work.

\newpage
\bibliography{references}

\appendix

\end{document}